\documentclass[11pt,journal,compsoc,onecolumn]{IEEEtran}
 \usepackage{cite}

\ifCLASSINFOpdf
\else
\fi
\ifCLASSOPTIONcompsoc
 \usepackage[font=normalsize,labelfont=sf,textfont=sf]{subfig}
\else
 \usepackage[font=footnotesize]{subfig}
\fi
\ifCLASSOPTIONcompsoc
 \usepackage[font=normalsize,labelfont=sf,textfont=sf]{subfig}
\else
 \usepackage[font=footnotesize]{subfig}
\fi

\usepackage{color}
 \usepackage{amssymb}
\usepackage{amsfonts}
\usepackage{dsfont}
\usepackage{epsfig}

\title{Human Action Recognition with Deep Temporal Pyramids}
\author{Ahmed Mazari \ \ \ \ \ \ \ \ \ \ \ \ \ \ Hichem Sahbi \\ $ $ \\ {CNRS, LIP6 Lab, Sorbonne University, Paris}}

\begin{document}

\maketitle

\begin{abstract}

Deep convolutional neural networks (CNNs) are nowadays achieving significant leaps in different pattern recognition tasks including action recognition. Current CNNs are increasingly deeper, data-hungrier and this makes their success tributary of the abundance of labeled training data. CNNs also rely on max/average pooling which reduces dimensionality of output layers and hence attenuates their sensitivity to the availability of labeled data. However, this process may dilute the information of upstream convolutional layers and thereby affect the discrimination power of the trained representations, especially when the learned categories are fine-grained.\\
  \indent In this paper, we introduce a novel hierarchical aggregation design, for final pooling, that controls granularity of the learned representations w.r.t the actual granularity of action categories. Our solution is based on a tree-structured temporal pyramid that aggregates outputs of CNNs at different levels. Top levels of this hierarchy are dedicated to coarse  categories while deep levels are more suitable to fine-grained ones. The design of our temporal pyramid is based on solving a constrained minimization problem whose solution corresponds to the distribution of weights of different representations in the temporal pyramid. Experiments conducted using the challenging UCF101 database show the relevance of our hierarchical design w.r.t other related methods.
 \end{abstract}


\section{Introduction}
\label{sec:intro}

\indent Video action recognition is a major challenge in computer vision which consists in assigning labels (also known as classes or categories) to  sequences of video frames. The challenge in action recognition stems from (i) the difficulty to learn mapping models that assign action categories to frames while being resilient to their acquisition conditions (namely occlusion, illumination, spatial-temporal resolution/scale/length, camera motion and velocity, truncation, background clutter, etc.) and also (ii) the hardness in hand-labeling large collections of training videos prior to build these mapping models.\\

\indent Existing action recognition techniques are usually based on machine learning~\cite{du2015hierarchical,sahbi2011context,ling2015,postadjian2017investigating,temporalpyramid,sahbi2002face,segment_net,chen2006human,wang2013directed,temporal_pyramid,sahbi2013cnrs,wang2015action,gowayyed2013histogram,sahbi2003coarse,superived_dic_action,multi_svm}; their general principle consists first in describing video frames using handcrafted or learned representations~\cite{boujemaa2001ikona,wang2012,sahbi2007kernel,fathi2008,napoleon20102d,wang2016action,xia2012,ferecatu2008telecomparistech,ali2010,murthy2013ordered,boujemaa2004visual,Le2011,tollari2008comparative,Matikainen2009} and then assigning these representations to action categories using variety of machine learning and inference techniques \cite{li2011superpixel,yu2010,mkl_action,poppe2010,sahbi2015imageclef,Iosifidis2013} including support vector machines~\cite{wu2012view,temporal_pyramid,sahbi2002coarse,wu2011action,pyramid_kernel,danafar2007action,wang2014bags} and deep networks~\cite{kin3d,pose,jiu2017nonlinear,baccouche2011,spresnet16,spresnetmulti17}. In particular, deep learning models are successful but their success is highly dependent on the availability of large collections of hand-labeled videos\footnote{that cover all the intra-class variability of action categories.} which are usually difficult to collect and expensive \cite{thiemert2006using,bourdis2011constrained,thiemert2005applying} -- even at reasonable frame rates  -- especially when handling fine-grained action categories. As a result, existing labeled training sets, for action recognition, are at least two orders of magnitude smaller compared to other neighboring tasks (such as the well studied image classification \cite{sahbi2010context,vo2014,sahbi2008context,oliveau2018,jiu2015semi}) while action recognition is intrinsically more challenging. Furthermore, training and fine-tuning these models, together with their hyper-parameters for the challenging task of action recognition, is known to be memory and time demanding even when using highly efficient GPU resources and reasonable size videos.\\

\noindent The increase in the discrimination power of the aforementioned convolutional networks (due to an increase in the number of their parameters) comes to the detriment of an increase of their sensitivity to the acquisition conditions especially on challenging datasets such as the UCF101. Hence, these networks become more data-hungry and more subject to over-fitting. Pooling based on average or max operators, also known as aggregation, attenuates such effect and makes it possible to reduce the sensitivity and hence enhances the resilience of these CNNs to the lack of training data and thereby to the acquisition conditions. However, pooling produces a downside effect: a loss in the discrimination power especially when videos belong to fine-grained action categories. Put differently, convolutional layers without pooling help discriminating fine-grained categories while pooling helps  discriminating coarse-grained categories; choosing a granularity for pooling is clearly a challenging task that {\it requires an appropriate design and this constitutes the main contribution of this work.} \\ 
\indent In this paper, we introduce a novel hierarchical aggregation design that balances the discrimination power of CNN outputs and their resilience to video acquisition conditions. Our solution is based on a temporal pyramid that aggregates the outputs of  CNNs at different levels, resulting into a hierarchical representation. Top levels of this hierarchy are dedicated to coarse action categories while deep levels are dedicated to fine-grained ones. The design principle of our temporal pyramid is based on solving a constrained minimization problem whose solution corresponds to the distribution of weights of different representations in the temporal pyramid. Experiments conducted, on action recognition, using the challenging UCF101 database show the substantial gain and the complementary aspect of our hierarchical design w.r.t other related methods~\cite{kin3d,pose}.

\def\S{{\cal S}}
\def\G{{\cal G}}
\def\V{{\cal V}}
\def\E{{\cal E}}
\def\F{{\cal F}}
\def\K{{\cal K}}
\def\N{{\cal N}}

\section{Proposed Method}

Considering a collection of videos $\S=\{\V_i\}_{i=1}^n$, with each one being a sequence of frames $\V_i=\{f_{i,t}\}_{t=1}^{T_i}$; in this paper, and unless explicitly mentioned, the symbol $i$ is omitted and $\V_i$, $f_{i,t}$ are simply rewritten as $\V$, $f_t$ respectively. As shown subsequently, frames in $\V$ are described using ``end-to-end'' trained network representations (see details in the subsequent sections). Without a loss of generality, we assume $T_i$ constant and simply denoted as $T$; otherwise frame sampling could be achieved to make $T_i$ constant.\\
\indent In what follows, we first present the branches used to build a representation at the frame-level. Then, we show how these frame-level representations are aggregated and combined at the video-level in order to achieve highly effective action recognition. Finally, we discuss our hierarchical aggregation design and mainly the learning of its parameters.
\subsection{Deep frame-wise representations}
In order to describe the visual content of a given video $\V$, we rely on a two-stream process; the latter provides a complete description of {\it appearance} and {\it motion}, based on \cite{pose,kin3d}, that characterizes the spatio-temporal aspects of moving objects and their interactions. The output of the appearance stream (denoted as $\{\psi_a(f_t)\}_t \subset \mathbb{R}^q$ with $q=2048$ in practice) is based on the deep residual network ({ResNet-152}) trained on ImageNet \cite{imagenet}. Besides the high performances reported in ImageNet classification~\cite{art_imagenet}, the particularity and the strength of this network resides in its skip connections which (i) reduce the sensitivity of the network to its architecture and (ii) reduce the effect of gradient collapse/explosion thereby making the optimization and fine-tuning of this network parameters (through stochastic gradient descent) effective and numerically more stable. The output of the motion stream (denoted as $\{\psi_m(f_t)\}_t$) is based either on 3D CNN \cite{kin3d} or 2D CNN \cite{pose};  the former is trained with normalized multi-frame  optical flows\footnote{Normalized means that the values of the optical flow  range between $0$ and $255$.} while the latter is trained with  heatmaps colorized at the video-level.

\subsection{Deep hierarchical aggregation}
Given a video $\V$, we introduce in this section an aggregation process that combines representations obtained at the frame-level of $\V$. A good aggregation design should tradeoff the global description of videos while capturing their details that distinguish possible fine-grained categories of actions. Hence, the design principle of our aggregation process is tree-structured and relies on a hierarchy of convolutional network representations. Without a loss of generality, we consider a binary hierarchy of $L$ levels, where ${\cal N}_{k,\ell}$ stands for the set of frames that belong to the $k$-th node and the $\ell$-th level. Top levels of this hierarchy provide {\it coarse} (long-term) video representations that capture global motion and appearance of actions while deep levels capture  {\it fine} (and timely-resolute) details of these actions, such as ``beginning'', ``middle'' and ``late'' aspects of actions, resulting into {\it coarse-to-fine} spatio-temporal representations. \\
\indent Each node  ${\cal N}_{k,\ell}$ is assigned an appearance representation, referred to as $\Psi_a^{k,\ell}(\V)$; this representation is  defined as  $\Psi_a^{k,\ell}(\V)  = \frac{1}{| {\cal N}_{k,\ell}|} \sum_{t \in {\cal N}_{k,\ell}}  \psi_a(f_t)$. Given a set of action categories ${\cal C}=\{1,\dots,C\}$;  we train multiple classifiers (denoted $\{g_c\}_{c \in {\cal C}}$) on top of the hierarchy of these representations. In practice, we use SVMs as classifiers whose kernels correspond to linear combinations of elementary kernels dedicated to $\{{\cal N}_{k,\ell}\}_{k,\ell}$. SVMs are suitable choices as they allow us to weight the impact of nodes in the hierarchy and put more emphasis on the most relevant granularity of the learned representations. Hence, depending on the granularity of action categories, SVMs will prefer top or deep layers of the hierarchy. \\

\noindent Considering a training set of videos $\{(\V_i,y_{ic})\}_i$ associated to an action category c, with $y_{ic} =+1$ if $\V_i$ belongs to the category $c$ and $y_{ic} =-1$ otherwise, the SVM associated to this action category $c$ is given by $g_c(\V) = \sum_{i} \alpha_i^c  y_{ic}  \K(\V,\V_i) + b_c$, here $b_c$ is a shift, $\{\alpha_i^c\}_i$ is a set of positive parameters and $\K$ is a positive semi-definite kernel; details about the setting of $\K$, as a part of SVM training and hierarchical aggregation design of our temporal pyramid, are given in the subsequent section.
\subsection{Coarse-to-fine hierarchical aggregation design} 
Let ${\cal N}=\cup_{k,\ell} \N_{k,\ell}$ be the union of all possible nodes (frame sets) in the hierarchy of {\it depth} up to $L$ levels and {\it width} up to $2^L$ nodes; in this section, we introduce our hierarchical aggregation design that allows us to combine multiple representations in $\cal N$. Our method is based on learning a convex combination of representations, and finds the ``optimal'' weights of this combination while training multi-class SVMs.  \\
\indent In what follows, unless explicitly mentioned, we write $\Psi^{k,\ell}_a(\V_i)$ for short as $\Psi^{k,\ell}(\V_i)$. We consider $\{\Psi^{k,\ell}(\V_i)\}_{i}^n$ as a training set of representations and $y_i \in \{1,\dots, C\}$ as the label (or category) of $\Psi^{k,\ell}(\V_i)$ taken from a well defined ground-truth; in practice $C=101$ (see experiments). Multi-class SVMs use the mapping $\Psi^{k,\ell}(\V)$ that takes a given video $\V$ from an input space into its representation space and find the unknown label of $\V$ as
\begin{equation}
\arg\max_{ c \in {\cal C}} g_{c}^{k,\ell}(\V),  
\end{equation} 
here $g_{c}^{k,\ell}(\V)=\langle w_c^{k,\ell}, \Psi^{k,\ell}(\V)\rangle + b_{k,\ell}$, with  $w_c^{k,\ell}$, $b_{k,\ell}$ being respectively hyperplane normal and bias associated to a given category $c \in {\cal C}$ and node  $(k,\ell)$. \\
\indent In order to combine different nodes in the hierarchy and hence design appropriate aggregation, we use multiple representation learning that generalizes the above SVM framework \cite{MKL}. Its main idea consists in finding a kernel $\K$ as a convex linear combination of positive semi-definite (p.s.d) elementary kernels $\{\K_{k,\ell}\}_{k,\ell}$ associated to $\{\N_{k,\ell}\}_{k,\ell}$. Thus, the kernel value between two videos $\V$, $\V'$ is defined as
\begin{equation} 
\K(\V,\V') =\sum_{k=1}^L \sum_{\ell=1}^{2^k} \beta_{k,\ell} \ \K_{k,\ell} (\V,\V'), 
\end{equation}
here $\beta_{k,\ell} \geq 0$, $\sum_{k,\ell} \beta_{k,\ell}=1$ and each kernel $\K_{k,\ell}$ operates using only the subset $\N_{k,\ell}$ (in practice,  $\K_{k,\ell}(\V,\V') =\langle \Psi^{k,\ell}(\V), \Psi^{k,\ell}(\V')\rangle$). Resulting from the closure of the p.s.d of $\{\K_{k,\ell}\}_{k,\ell}$ w.r.t the sum, the final kernel $\K$ will also be p.s.d. Hence, using a primal SVM formulation, we predict the unknown category of a given video $\V$ as
 $\arg\max_{c\in {\cal C}} g_c(\V)$, with $g_c(\V)=\sum_{k,\ell} \beta_{k,\ell} \langle w_c^{k,\ell},\Psi^{k,\ell}(\V)\rangle+b_c$ and $b_c$, $\{w_c^{k,\ell}\}_{k,\ell}$ being respectively the bias and the hyperplane normals associated to a given class $c$ for different nodes. We choose the parameters $\beta=\{\beta_{k,\ell}\}_{k,\ell}$, $b=\{b_c\}$ and $w=\{w_c^{k,\ell}\}_{k,\ell}$ by solving the following constrained minimization problem
\begin{equation}
  \begin{array}{ll}
    \displaystyle 
    \min_{\beta,w,b,\xi} &  \displaystyle  \frac{1}{2} \sum_{k,\ell} \sum_c \beta_{k,\ell} \langle w_c^{k,\ell}, w_c^{k,\ell}\rangle  + \sum_{j=1}^{n}\xi_j \\
    &  \\
   \textrm{s.t.}              &   \displaystyle  \xi_j = \max_{c'\in {\cal C}\backslash c} l(g_c(\V_j)-g_{c^{'}}(\V_j)), 
\end{array}
                       \end{equation} 
                       here $c \in {\cal C}$ is the actual label of $\V_j$,  $\xi=\{\xi_j\}_j$ acts as a softmax and $l(.)$ is a convex loss function.  As this problem is not convex w.r.t the training parameters $\beta,w,b,\xi$ taken jointly and convex when taken separately, an EM-like iterative optimization procedure can be used: first, $\beta$ is fixed and the above problem is solved w.r.t $w,b,\xi$ using quadratic programming, then  $w,b,\xi$ are fixed and the resulting problem is solved w.r.t $\beta$ using linear programming. This iterative process stops when the values of all these parameters remain unchanged (from one iteration to another) or when it reaches a maximum number of iterations (see for instance~\cite{MKL}).
        
\section{Experiments}
\begin{figure} 
     \centering
\resizebox{0.95\columnwidth}{!}{ 
  \includegraphics[width=10cm]{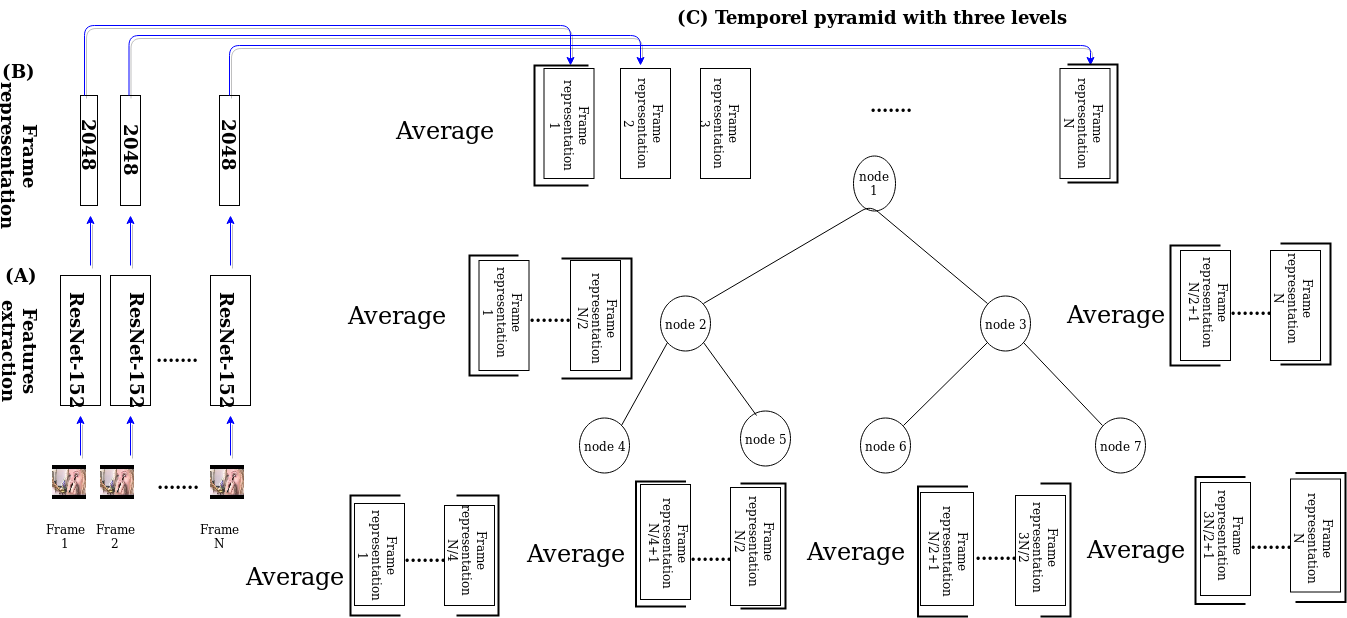}

    }
    \caption{This figure shows steps (A, B and C) of deep hierarchical aggregation that controls (via $\{\beta_{k,\ell}\}_{k,\ell}$) the granularity of the learned representations. {\bf Better to zoom the PDF.}}
    \label{fig:pyramid} 
\end{figure}

\begin{figure}
 \centering
 \includegraphics[width=8cm]{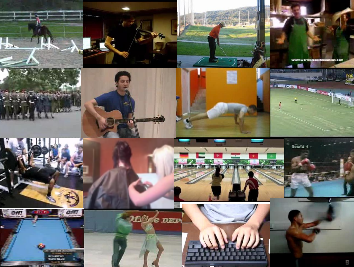}
 \caption{\small Sample of classes from UCF-101 dataset. From top-left to bottom-right, classes are: riding horse, playing violin, golf swing, pizza tossing, military parade, playing guitar, pushups, soccer penalty, bench press, haircut, bowling, punch, billiard, ice-dancing, typing.}
 \label{fig:ucf_101_classes}
\end{figure}

In this section, we evaluate the performance of action classification using the challenging UCF101 database \cite{ucf}. This dataset includes 13320 video shots taken from various actions belonging to  101 categories. These videos have diverse contents and were taken under extremely challenging and uncontrolled conditions, with many viewpoint changes (see examples of video frames in Fig.~\ref{fig:ucf_101_classes}). Each video is processed in order to extract its underlying CNN representations at the frame-level, followed by their hierarchical aggregation at the video-level as discussed in section 2;  Fig.~\ref{fig:pyramid} is an illustration of the whole video representation process. 
\subsection{Setting and evaluation protocol}
The purpose of our  evaluation is to show the performance of the hierarchical aggregation design of our temporal pyramid (TP) compared to different coarse and  fine aggregations as well as other baselines. We also extend the comparison of action classification against reported results in the related work.  \\
\noindent We plugged our temporal pyramid into support vector classifiers in order to evaluate their performances. Again the targeted task is action classification (a.k.a recognition); given a video shot described with a temporal pyramid, the goal is to predict which action (class) is present into that shot. For this purpose, we trained a one-vs.-all SVM classifier for each class; we use the train-test split2 evaluation protocol (suggested in  \cite{ucf})  in order to compare the performance of our method against the related work under exactly the same conditions. We repeat this training and testing process through different classes and we take the average accuracy over all the classes of actions. 
\subsection{Performance and comparison}
\noindent {\bf Baselines.} We first show a comparison of action recognition performance, using our temporal pyramid, against two baselines:  global average pooling and also spectrograms; the {\it former}  produces a global representation that averages all the frame descriptions while the {\it latter} keeps all the frame representations and concatenate them prior to their classification. Note that these two comparative baselines  are interesting as they correspond  to two extreme cases of our hierarchy, namely the root and the leaf levels; in particular, the spectrogram (of a video $\V$ with $T$ frames) is  obtained when the number of leaf nodes, in the temporal pyramid, is exactly equal  to $T$.  We also consider as baselines: level-wise representations of our temporal pyramid.  Early observations, reported in Table.~\ref{table1}, show that our hierarchical representation design makes it possible select the best configuration (combination) of level representations in order to improve the classification accuracy; indeed, the results show a clear gain when using this design compared to all the other levels when taken separately as well as spectrograms. This gain results from the match between the granularity of the learned representations  in the temporal pyramid and the actual granularities of  action categories.\\
\begin{table}
  \begin{center}
\resizebox{0.69\columnwidth}{!}{
\textcolor{black}{\begin{tabular}{c||c}
      Setting & action recognition performance on UCF101 \\
      \hline
      \hline 
       Temporal pyramid (level 1) &   66.15\%    \\
        Temporal pyramid (level 2) &  66.74\%    \\
       Temporal pyramid (level 3)  &  67.14\%    \\
       Temporal pyramid (level 4) &  67.41\%   \\
      Temporal pyramid (level 5) &   67.45\%   \\
                    Temporal pyramid (level 6) &  67.47\%  \\
                    \hline 
                    Temporal pyramid + Multiple Rep   & {\bf 68.58\%} \\
                    \hline 
                    Spectrograms   & 64.41\%                                                    
                \end{tabular}}
              }
    \caption{\small Performances (on split 2 of UCF101) for level-wise and with multiple representation learning (referred to as ''Multiple Rep'', see also~\cite{MKL_alignement}) as described in section 2.3. As already described, level 1 corresponds to the global average pooling.}  \label{table1}
\end{center}
  \end{table}

\noindent {\bf Comparison w.r.t the related work.}  We also compared the classification performances and the complementary aspect of our temporal pyramid design against related work including  \cite{pose} and \cite{kin3d}.  The method in  \cite{pose} is based on colorized heatmaps, as a variant of the global average pooling baseline; the latter corresponds to timely-stamped and averaged frame-wise probability distributions of human keypoints.  From the results in Table.~\ref{table2}, our hierarchical design, brings a substantial gain of at least 12 points w.r.t  colorized heatmaps.  The method in \cite{kin3d} is based on 3D CNNs including two-streams; one  for motion and another one for appearance. While these two-streams are highly effective, their combination with our temporal pyramid, through a simple late fusion, brings a noticeable gain. We also observe the same behavior on all the combinations -- of this two stream CNN with the other baselines;   despite the fact that bridging the last few percentage gap is very challenging,  for each setting  our temporal pyramid succeeds in improving the performances. 
  \begin{table}[h]  \centering
 \resizebox{.75\columnwidth}{!}{\begin{tabular}{c||c}
      Method & action recognition performances   \\
      \hline
      \hline 
      col. heatM   \cite{pose}                 &   64.38\%  \\
      col. heatM  \cite{pose}  {\bf +TP}  & {\bf 77.34\%} \\
      \hline
      Spect   & 64.41\%   \\
      Spect  \textcolor{black}{{\bf +TP}}  & \textcolor{black}{68.40\%} \\
      Spect + col.  heatM \cite{pose}  &  66.87\% \\
      Spect + col.  heatM\cite{pose} \textcolor{black}{{\bf +TP}}  &  \textcolor{black}{\bf 74.65\%} \\
       \hline 
      3D 2-stream (motion) \cite{kin3d}                        &  96.41\%   \\
      3D 2-stream (appearance) \cite{kin3d}                     & 95.60\%  \\
      3D 2-stream (combined) \cite{kin3d}                       & {97.94\%}  \\
      3D 2-stream (motion) \cite{kin3d} \textcolor{black}{{\bf +TP}}    &  {97.50\%}   \\
      3D 2-stream (appearance) \cite{kin3d} \textcolor{black}{{\bf +TP}}    & \textcolor{black}{95.77\%}  \\
      3D 2-stream (combined) \cite{kin3d} \textcolor{black}{{\bf +TP}}    & \textcolor{black}{97.94\%}  \\
      \hline       
      3D 2-stream (motion) \cite{kin3d} + col. heatM  \cite{pose} &  94.89\%    \\
      3D 2-stream (appearance) \cite{kin3d} + col. heatM \cite{pose} & 94.32\% \\
      3D 2-stream (combined) \cite{kin3d} + col. heatM  \cite{pose} & 97.02\%  \\
      3D 2-stream (motion) \cite{kin3d} + col.  heatM \cite{pose} \textcolor{black}{{\bf +TP}}    & \textcolor{black}{95.70\%}  \\             
      3D 2-stream (appearance) \cite{kin3d} + col.  heatM \cite{pose} \textcolor{black}{{\bf +TP}} & \textcolor{black}{94.60\%}  \\             
      3D 2-stream (combined) \cite{kin3d} + col.  heatM  \cite{pose} \textcolor{black}{{\bf +TP}}   & \textcolor{black}{\bf 97.56\%}  \\             
      \hline       
      3D 2-stream (motion) \cite{kin3d} + spect         & 95.64\%   \\
      3D 2-stream (appearance) \cite{kin3d} + spect     & 94.72\%  \\
      3D 2-stream (combined) \cite{kin3d} + spect       & 97.70\%  \\
      3D 2-stream (motion) \cite{kin3d} + spect \textcolor{black}{{\bf +TP}}      & \textcolor{black}{95.77\%}   \\
      3D 2-stream (appearance) \cite{kin3d} + spect \textcolor{black}{{\bf +TP}}  & \textcolor{black}{94.95\%}   \\
      3D 2-stream (combined) \cite{kin3d} + spect \textcolor{black}{{\bf +TP}}    & \textcolor{black}{\bf 97.74\%}   \\
      \hline       
      3D 2-stream (motion) \cite{kin3d} + col. heatM  \cite{pose} + spect    & 95.12\% \\
      3D 2-stream (appearance) \cite{kin3d} + col. heatM  \cite{pose} + spect  & 94.70\% \\
      3D 2-stream (combined) \cite{kin3d} + col. heatM  \cite{pose} + spect   & 97.32\% \\
      3D 2-stream (motion) \cite{kin3d}+ col. heatM  \cite{pose} + spect  \textcolor{black}{{\bf +TP}}   & \textcolor{black}{96.35\%}  \\
      3D 2-stream (appearance) \cite{kin3d}+ col. heatM  \cite{pose} + spect  \textcolor{black}{{\bf +TP}}  & \textcolor{black}{95.10\%}   \\
      3D 2-stream (combined) \cite{kin3d}+ col. heatM \cite{pose}  + spect  \textcolor{black}{{\bf +TP}}  & \textcolor{black}{\bf 97.51\%}                        
    \end{tabular}}
\caption{\small Comparison w.r.t the related work (on split2 of UCF101 dataset); in this table: ``heatM'' stands for colorized heatmaps, ``spect'' for spectrograms and TP for temporal pyramid + multiple representation. We observe a clear gain (highlighted in bold) when TP is used and combined w.r.t the related work.} \label{table2}
\end{table}
\section{Conclusion} 
We introduced in this paper an action recognition method based on convolutional neural networks and a novel hierarchical aggregation design. The latter defines pooling operations at different granularities and makes it possible to fit the actual granularity of action categories resulting into a clear gain in performance compared to global average pooling and also spectrograms. Our method is based on solving a constrained minimization problem whose solution corresponds to the level-wise weight distributions which also maximize performances. Comparison, using the challenging UCF101 dataset,  shows the validity and the complementary aspect of our method with respect to the related work. 
As a future work  we are currently investigating the application of our hierarchical aggregation to activity recognition, on longer duration video datasets, and this requires  deeper temporal pyramids. 

\bibliographystyle{IEEEtran}
\bibliography{references}

\begin{thebibliography}{10}
\providecommand{\url}[1]{#1}
\csname url@samestyle\endcsname
\providecommand{\newblock}{\relax}
\providecommand{\bibinfo}[2]{#2}
\providecommand{\BIBentrySTDinterwordspacing}{\spaceskip=0pt\relax}
\providecommand{\BIBentryALTinterwordstretchfactor}{4}
\providecommand{\BIBentryALTinterwordspacing}{\spaceskip=\fontdimen2\font plus
\BIBentryALTinterwordstretchfactor\fontdimen3\font minus
  \fontdimen4\font\relax}
\providecommand{\BIBforeignlanguage}[2]{{%
\expandafter\ifx\csname l@#1\endcsname\relax
\typeout{** WARNING: IEEEtran.bst: No hyphenation pattern has been}%
\typeout{** loaded for the language `#1'. Using the pattern for}%
\typeout{** the default language instead.}%
\else
\language=\csname l@#1\endcsname
\fi
#2}}
\providecommand{\BIBdecl}{\relax}
\BIBdecl

\bibitem{du2015hierarchical}
Y.~Du, W.~Wang, and L.~Wang, ``Hierarchical recurrent neural network for
  skeleton based action recognition,'' in \emph{Proceedings of the IEEE
  conference on computer vision and pattern recognition}, 2015, pp. 1110--1118.

\bibitem{sahbi2011context}
H.~Sahbi, J.-Y. Audibert, and R.~Keriven, ``Context-dependent kernels for
  object classification,'' \emph{IEEE transactions on pattern analysis and
  machine intelligence}, vol.~33, no.~4, pp. 699--708, 2011.

\bibitem{ling2015}
L.~Wang, ``kernel machines for video action recognition,'' in \emph{PhD thesis,
  Telecom ParisTech, Paris-Saclay University}, 2015.

\bibitem{postadjian2017investigating}
T.~Postadjian, A.~Le~Bris, H.~Sahbi, and C.~Mallet, ``Investigating the
  potential of deep neural networks for large-scale classification of very high
  resolution satellite images,'' \emph{ISPRS Annals}, vol.~4, pp. 183--190,
  2017.

\bibitem{temporalpyramid}
H.~Pirsiavash and D.~Ramanan, ``Detecting activities of daily living in
  first-person camera views,'' in \emph{Computer Vision and Pattern Recognition
  (CVPR), 2012 IEEE Conference on}.\hskip 1em plus 0.5em minus 0.4em\relax
  IEEE, 2012, pp. 2847--2854.

\bibitem{sahbi2002face}
H.~Sahbi, D.~Geman, and N.~Boujemaa, ``Face detection using coarse-to-fine
  support vector classifiers,'' in \emph{Image Processing. 2002. Proceedings.
  2002 International Conference on}, vol.~3.\hskip 1em plus 0.5em minus
  0.4em\relax IEEE, 2002, pp. 925--928.

\bibitem{segment_net}
L.~Wang, Y.~Xiong, Z.~Wang, Y.~Qiao, D.~Lin, X.~Tang, and L.~V. Gool,
  ``Temporal segment networks: Towards good practices for deep action
  recognition,'' \emph{ECCV}, 2016.

\bibitem{chen2006human}
H.-S. Chen, H.-T. Chen, Y.-W. Chen, and S.-Y. Lee, ``Human action recognition
  using star skeleton,'' in \emph{Proceedings of the 4th ACM international
  workshop on Video surveillance and sensor networks}.\hskip 1em plus 0.5em
  minus 0.4em\relax ACM, 2006, pp. 171--178.

\bibitem{wang2013directed}
L.~Wang and H.~Sahbi, ``Directed acyclic graph kernels for action
  recognition,'' in \emph{Proceedings of the IEEE International Conference on
  Computer Vision}, 2013, pp. 3168--3175.

\bibitem{temporal_pyramid}
D.~Xu and S.-F. Chang, ``Visual event recognition in news video using kernel
  methods with multi-level temporal alignment,'' \emph{CVPR}, 2007.

\bibitem{sahbi2013cnrs}
H.~Sahbi, ``Cnrs-telecom paristech at imageclef 2013 scalable concept image
  annotation task: Winning annotations with context dependent svms.'' in
  \emph{CLEF (Working Notes)}, 2013.

\bibitem{wang2015action}
L.~Wang, Y.~Qiao, and X.~Tang, ``Action recognition with trajectory-pooled
  deep-convolutional descriptors,'' in \emph{Proceedings of the IEEE conference
  on computer vision and pattern recognition}, 2015, pp. 4305--4314.

\bibitem{gowayyed2013histogram}
M.~A. Gowayyed, M.~Torki, M.~E. Hussein, and M.~El-Saban, ``Histogram of
  oriented displacements (hod): Describing trajectories of human joints for
  action recognition,'' in \emph{Twenty-Third International Joint Conference on
  Artificial Intelligence}, 2013.

\bibitem{sahbi2003coarse}
H.~Sahbi, ``Coarse-to-fine support vector machines for hierarchical face
  detection,'' Ph.D. dissertation, PhD thesis, Versailles University, 2003.

\bibitem{superived_dic_action}
H.~Wang, C.~Yuan, W.~Hu, and C.~Sun, ``Supervised class-specific dictionary
  learning for sparse modeling in action recognition,'' \emph{Pattern
  Recognition}, vol.~45, no.~11, pp. 3902--3911, 2012.

\bibitem{multi_svm}
C.~Schuldt, I.~Laptev, and B.~Caputo, ``Recognizing human actions: a local svm
  approach,'' in \emph{Pattern Recognition, 2004. ICPR 2004. Proceedings of the
  17th International Conference on}, vol.~3.\hskip 1em plus 0.5em minus
  0.4em\relax IEEE, 2004, pp. 32--36.

\bibitem{boujemaa2001ikona}
N.~Boujemaa, J.~Fauqueur, M.~Ferecatu, F.~Fleuret, V.~Gouet, B.~Saux, and
  H.~Sahbi, ``Ikona: Interactive generic and specific image retrieval,'' in
  \emph{Proceedings of the International workshop on Multimedia Content-Based
  Indexing and Retrieval (MMCBIR?2001)}, 2001, pp. 25--29.

\bibitem{wang2012}
J.~Wang, Z.~Liu, Y.~Wu, and J.~Yuan, ``Mining actionlet ensemble for action
  recognition with depth cameras,'' \emph{In Computer Vision and Pattern
  Recognition (CVPR), 2012 IEEE Conference on (pp. 1290-1297)}, 2012.

\bibitem{sahbi2007kernel}
H.~Sahbi, ``Kernel pca for similarity invariant shape recognition,''
  \emph{Neurocomputing}, vol.~70, no. 16-18, pp. 3034--3045, 2007.

\bibitem{fathi2008}
A.~Fathi and G.~Mori, ``Action recognition by learning mid-level motion
  features.'' \emph{In Computer Vision and Pattern Recognition, 2008. CVPR
  2008. IEEE Conference on (pp. 1-8). IEEE.}, 2008.

\bibitem{napoleon20102d}
T.~Napol{\'e}on and H.~Sahbi, ``From 2d silhouettes to 3d object retrieval:
  contributions and benchmarking,'' \emph{Journal on Image and Video
  Processing}, vol. 2010, p.~1, 2010.

\bibitem{wang2016action}
P.~Wang, Z.~Li, Y.~Hou, and W.~Li, ``Action recognition based on joint
  trajectory maps using convolutional neural networks,'' in \emph{Proceedings
  of the 24th ACM international conference on Multimedia}.\hskip 1em plus 0.5em
  minus 0.4em\relax ACM, 2016, pp. 102--106.

\bibitem{xia2012}
L.~Xia, C.~Chen, and J.~K. Aggarwal, ``View invariant human action recognition
  using histograms of 3d joints.'' \emph{In Computer vision and pattern
  recognition workshops (CVPRW), 2012 IEEE computer society conference on (pp.
  20-27). IEEE.}, 2012.

\bibitem{ferecatu2008telecomparistech}
M.~Ferecatu and H.~Sahbi, ``Telecomparistech at imageclefphoto 2008: Bi-modal
  text and image retrieval with diversity enhancement.'' in \emph{CLEF (Working
  Notes)}, 2008.

\bibitem{ali2010}
S.~Ali and M.~Shah, ``Human action recognition in videos using kinematic
  features and multiple instance learning.'' \emph{IEEE transactions on pattern
  analysis and machine intelligence, 32(2), 288-303.}, 2010.

\bibitem{murthy2013ordered}
O.~Murthy and R.~Goecke, ``Ordered trajectories for large scale human action
  recognition,'' in \emph{Proceedings of the IEEE international conference on
  computer vision workshops}, 2013, pp. 412--419.

\bibitem{boujemaa2004visual}
N.~Boujemaa, F.~Fleuret, V.~Gouet, and H.~Sahbi, ``Visual content extraction
  for automatic semantic annotation of video news,'' in \emph{the proceedings
  of the SPIE Conference, San Jose, CA}, vol.~6, 2004.

\bibitem{Le2011}
Q.~V. Le, W.~Zou, S.~Y. Yeung, and A.~Y. Ng, ``Learning hierarchical invariant
  spatio-temporal features for action recognition with independent subspace
  analysis.'' \emph{In Computer Vision and Pattern Recognition (CVPR), 2011
  IEEE Conference on (pp. 3361-3368). IEEE.}, 2011.

\bibitem{tollari2008comparative}
S.~Tollari, P.~Mulhem, M.~Ferecatu, H.~Glotin, M.~Detyniecki, P.~Gallinari,
  H.~Sahbi, and Z.-Q. Zhao, ``A comparative study of diversity methods for
  hybrid text and image retrieval approaches,'' in \emph{Workshop of the
  Cross-Language Evaluation Forum for European Languages}.\hskip 1em plus 0.5em
  minus 0.4em\relax Springer, 2008, pp. 585--592.

\bibitem{Matikainen2009}
P.~Matikainen, M.~Hebert, and R.~Sukthankar, ``Trajectons: Action recognition
  through the motion analysis of tracked features.'' \emph{In Computer Vision
  Workshops (ICCV Workshops), 2009 IEEE 12th International Conference on (pp.
  514-521). IEEE.}, 2009.

\bibitem{li2011superpixel}
X.~Li and H.~Sahbi, ``Superpixel-based object class segmentation using
  conditional random fields,'' in \emph{Acoustics, Speech and Signal Processing
  (ICASSP), 2011 IEEE International Conference on}.\hskip 1em plus 0.5em minus
  0.4em\relax IEEE, 2011, pp. 1101--1104.

\bibitem{yu2010}
T.~Yu, T.~Kim, and R.~Cipolla, ``Real-time action recognition by spatiotemporal
  semantic and structural forests.'' \emph{In BMVC (Vol. 2, No. 5, p. 6).},
  2010.

\bibitem{mkl_action}
L.~Chen, L.~Duan, and D.~Xu, ``Event recognition in videos by learning from
  heterogeneous web sources,'' in \emph{Proceedings of the IEEE Conference on
  Computer Vision and Pattern Recognition}, 2013, pp. 2666--2673.

\bibitem{poppe2010}
R.~Poppe, ``A survey on vision-based human action recognition.'' \emph{Image
  and vision computing, 28(6), 976-990.}, 2010.

\bibitem{sahbi2015imageclef}
H.~Sahbi, ``Imageclef annotation with explicit context-aware kernel maps,''
  \emph{International Journal of Multimedia Information Retrieval}, vol.~4,
  no.~2, pp. 113--128, 2015.

\bibitem{Iosifidis2013}
A.~Iosifidis, A.~Tefas, and I.~Pitas, ``Minimum class variance extreme learning
  machine for human action recognition.'' \emph{IEEE Transactions on Circuits
  and Systems for Video Technology, 23(11), 1968-1979.}, 2013.

\bibitem{wu2012view}
X.~Wu and Y.~Jia, ``View-invariant action recognition using latent kernelized
  structural svm,'' in \emph{European conference on computer vision}.\hskip 1em
  plus 0.5em minus 0.4em\relax Springer, 2012, pp. 411--424.

\bibitem{sahbi2002coarse}
H.~Sahbi and N.~Boujemaa, ``Coarse-to-fine support vector classifiers for face
  detection,'' in \emph{null}.\hskip 1em plus 0.5em minus 0.4em\relax IEEE,
  2002, p. 30359.

\bibitem{wu2011action}
S.~Wu, O.~Oreifej, and M.~Shah, ``Action recognition in videos acquired by a
  moving camera using motion decomposition of lagrangian particle
  trajectories,'' in \emph{2011 International conference on computer
  vision}.\hskip 1em plus 0.5em minus 0.4em\relax IEEE, 2011, pp. 1419--1426.

\bibitem{pyramid_kernel}
K.~Grauman and T.~Darrell, ``The pyramid match kernel: Efficient learning with
  sets of features,'' \emph{JMLR}, 2007.

\bibitem{danafar2007action}
S.~Danafar and N.~Gheissari, ``Action recognition for surveillance applications
  using optic flow and svm,'' in \emph{Asian Conference on Computer
  Vision}.\hskip 1em plus 0.5em minus 0.4em\relax Springer, 2007, pp. 457--466.

\bibitem{wang2014bags}
L.~Wang and H.~Sahbi, ``Bags-of-daglets for action recognition,'' in
  \emph{Image Processing (ICIP), 2014 IEEE International Conference on}.\hskip
  1em plus 0.5em minus 0.4em\relax IEEE, 2014, pp. 1550--1554.

\bibitem{kin3d}
J.~Carreira and A.~Zisserman, ``Quo vadis, action recognition? a new model and
  the kinetics dataset,'' in \emph{Computer Vision and Pattern Recognition
  (CVPR), 2017 IEEE Conference on}.\hskip 1em plus 0.5em minus 0.4em\relax
  IEEE, 2017, pp. 4724--4733.

\bibitem{pose}
V.~Choutas, P.~Weinzaepfel, J.~Revaud, and C.~Schmid, ``Potion: Pose motion
  representation for action recognition,'' \emph{CVPR}, 2018.

\bibitem{jiu2017nonlinear}
M.~Jiu and H.~Sahbi, ``Nonlinear deep kernel learning for image annotation,''
  \emph{IEEE Transactions on Image Processing}, vol.~26, no.~4, pp. 1820--1832,
  2017.

\bibitem{baccouche2011}
M.~Baccouche, F.~Mamalet, C.~Wolf, C.~Garcia, and A.~Baskurt, ``Sequential deep
  learning for human action recognition.'' \emph{In International Workshop on
  Human Behavior Understanding (pp. 29-39). Springer, Berlin, Heidelberg},
  2011.

\bibitem{spresnet16}
C.~Feichtenhofer, A.~Pinz, and R.~Wildes, ``Spatiotemporal residual networks
  for video action recognition,'' in \emph{Advances in neural information
  processing systems}, 2016, pp. 3468--3476.

\bibitem{spresnetmulti17}
C.~Feichtenhofer, A.~Pinz, and R.~P. Wildes, ``Spatiotemporal multiplier
  networks for video action recognition,'' in \emph{2017 IEEE Conference on
  Computer Vision and Pattern Recognition (CVPR)}.\hskip 1em plus 0.5em minus
  0.4em\relax IEEE, 2017, pp. 7445--7454.

\bibitem{thiemert2006using}
S.~Thiemert, H.~Sahbi, and M.~Steinebach, ``Using entropy for image and video
  authentication watermarks,'' in \emph{Security, Steganography, and
  Watermarking of Multimedia Contents VIII}, vol. 6072.\hskip 1em plus 0.5em
  minus 0.4em\relax International Society for Optics and Photonics, 2006, p.
  607218.

\bibitem{bourdis2011constrained}
N.~Bourdis, M.~Denis, and H.~Sahbi, ``Constrained optical flow for aerial image
  change detection,'' in \emph{2011 IEEE International Geoscience and Remote
  Sensing Symposium (IGARSS)}, 2011, pp. 4176--4179.

\bibitem{thiemert2005applying}
S.~Thiemert, H.~Sahbi, and M.~Steinebach, ``Applying interest operators in
  semi-fragile video watermarking,'' in \emph{Security, Steganography, and
  Watermarking of Multimedia Contents VII}, vol. 5681.\hskip 1em plus 0.5em
  minus 0.4em\relax International Society for Optics and Photonics, 2005, pp.
  353--363.

\bibitem{sahbi2010context}
H.~Sahbi and X.~Li, ``Context-based support vector machines for interconnected
  image annotation,'' in \emph{Asian Conference on Computer Vision}.\hskip 1em
  plus 0.5em minus 0.4em\relax Springer, 2010, pp. 214--227.

\bibitem{vo2014}
P.~Vo, ``Transductive inference and kernel methods for image annotation,'' in
  \emph{PhD thesis, Telecom ParisTech, Paris-Saclay University}, 2014.

\bibitem{sahbi2008context}
H.~Sahbi, J.-Y. Audibert, J.~Rabarisoa, and R.~Keriven, ``Context-dependent
  kernel design for object matching and recognition,'' in \emph{CVPR}, 2008,
  pp. 1--8.

\bibitem{oliveau2018}
Q.~Oliveau, ``Learning models for object and ship category recognition,'' in
  \emph{PhD thesis, Telecom ParisTech, Paris-Saclay University}, 2018.

\bibitem{jiu2015semi}
M.~Jiu and H.~Sahbi, ``Semi supervised deep kernel design for image
  annotation,'' in \emph{Acoustics, Speech and Signal Processing (ICASSP), 2015
  IEEE International Conference on}.\hskip 1em plus 0.5em minus 0.4em\relax
  IEEE, 2015, pp. 1156--1160.

\bibitem{imagenet}
J.~Deng, W.~Dong, R.~Socher, L.-J. Li, K.~Li, and L.~Fei-Fei, ``Imagenet: A
  large-scale hierarchical image database,'' \emph{IEEE Computer Vision and
  Pattern Recognition (CVPR)}, 2009.

\bibitem{art_imagenet}
B.~Zoph, V.~Vasudevan, J.~Shlens, and Q.-V. Le, ``Learning transferable
  architectures for scalable image recognition,'' \emph{CVPR}, 2018.

\bibitem{MKL}
M.~Gönen and E.~Alpaydın, ``Multiple kernel learning algorithms,''
  \emph{JMLR}, 2011.

\bibitem{ucf}
K.~Soomro, A.-R. Zamir, and M.~Shah, ``Ucf101: A dataset of 101 human action
  classes from videos in the wild,'' \emph{CRCV-TR-12-01}, 2012.

\bibitem{MKL_alignement}
C.cortes, M.~Mohri, and A.~Rostamizadeh, ``Algorithms for learning kernels
  based on centered alignement,'' \emph{J.Mach.Learn, Rev.}, vol.~13, pp.
  795--828, 2012.

\end{thebibliography}

\end{document}